\def\year{2021}\relax
\documentclass[letterpaper]{article} 
\usepackage{aaai21}  
\usepackage{times}  
\usepackage{helvet} 
\usepackage{courier}  
\usepackage[hyphens]{url}  
\usepackage{graphicx} 
\usepackage{natbib}  
\usepackage{caption} 
\usepackage{amsfonts}
\usepackage{url}
\usepackage{xcolor}
\usepackage{multirow}
\usepackage{amsbsy}
\usepackage{tabularx, booktabs, caption, makecell}
\usepackage{siunitx}

\title{A Graph Attention Based Approach for Trajectory Prediction \\ in Multi-agent Sports Games}
\author{
    Ding Ding,\textsuperscript{\rm 1}\thanks{This work was done when Ding Ding was a postdoctoral researcher at the George Washington University.}
    H. Howie Huang\textsuperscript{\rm 2}
    \\
}
\affiliations{
    \textsuperscript{\rm 1}Tencent America, USA \\
    \textsuperscript{\rm 2}The George Washington University, USA\\
    ddding@tencent.com, howie@gwu.edu
}

\begin{document}

\maketitle

\begin{abstract}
This work investigates the problem of multi-agents trajectory prediction. Prior approaches lack of capability of capturing fine-grained dependencies among coordinated agents. In this paper, we propose a spatial-temporal trajectory prediction approach that is able to learn the strategy of a team with multiple coordinated agents. In particular, we use graph-based attention model to learn the dependency of the agents. In addition, instead of utilizing the recurrent networks (e.g., VRNN, LSTM), our method uses a Temporal Convolutional Network (TCN) as the sequential model to support long effective history and provide important features such as parallelism and stable gradients.  We demonstrate the validation and effectiveness of our approach on two different sports game datasets: basketball and soccer datasets. The result shows that compared to  related approaches, our model that infers the dependency of players yields substantially improved performance. Code is available at \emph{\url{https://github.com/iHeartGraph/predict}}
\end{abstract}

\section{Introduction}\label{sec:Introduction}

Multi-agents prediction has been attracting great attention from both industry and academia researchers, due to the practical desire to deal with complex problems. In most of the real-world scenarios, multiple agents are coordinated in different roles but aim at achieving the same goal. 
For example, in a war, an agent wants to recognize a person who is coming over whether is a hostile enemy or a common civilian, and hence the agent and collaborators are able to execute some reactions. In a basketball game, an agent in the defensive team wants to understand the future trajectories of the offensive players, so that she and her teammates can make the optimized defense actions. 
It is an open challenge to design a sequential model to capture such complex behavior of multiple agents that is highly coordinated and non-deterministic. 

\begin{figure}[t]
  \centering
  \includegraphics[scale=0.17,trim={50 150 0 100},clip]{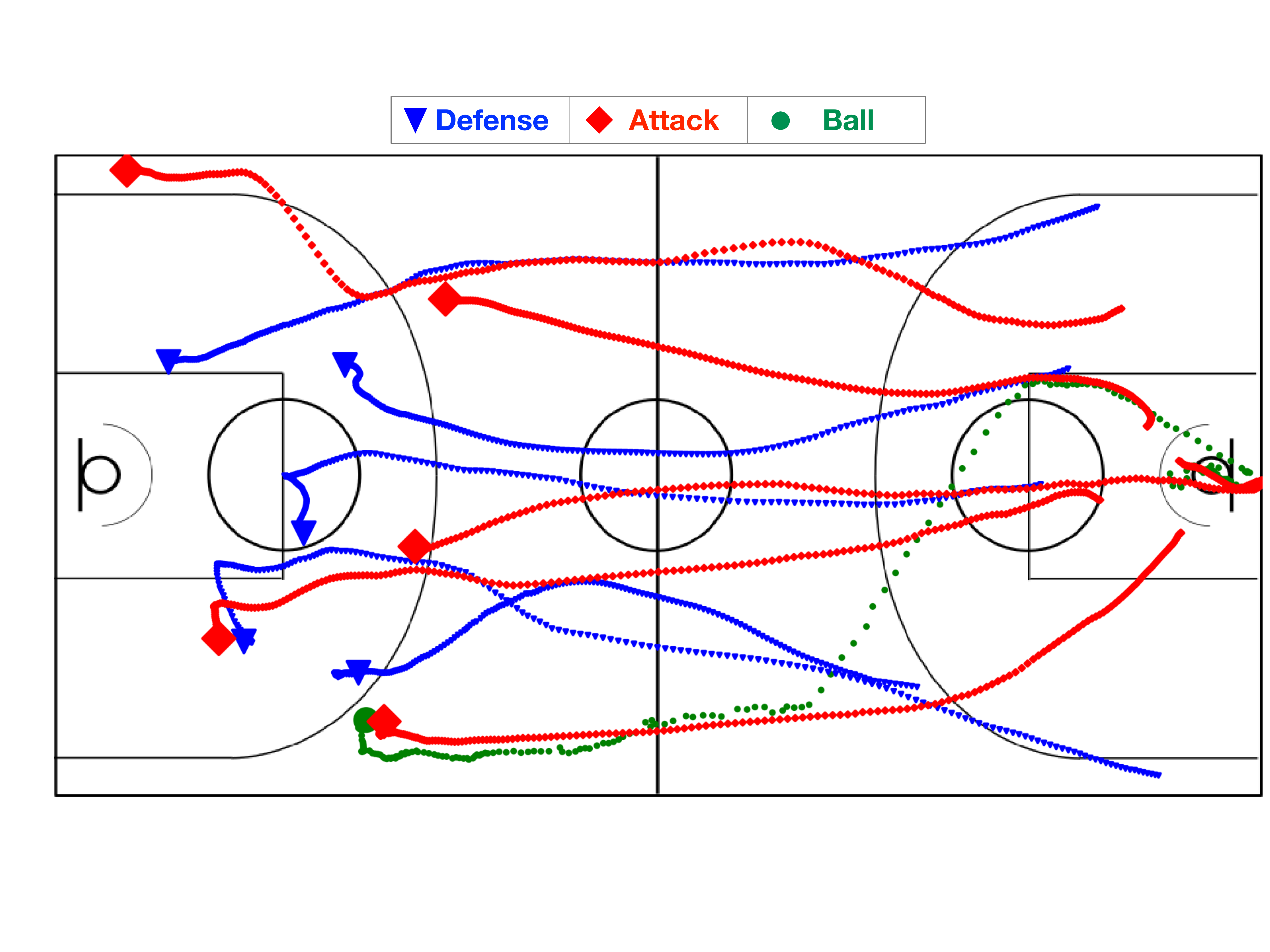}
  \caption{\textbf{Actual basketball game trajectories.} The trajectory data is from a Toronto Raptors NBA game in 2015, which contains offense (red), defense (blue), and ball (green) 2D overhead-view trajectories. \label{fig:basketballtrajectory}}
\end{figure}
In this paper, we focus on solving a multiple-agents context-aware prediction problem --- forecasting trajectories in sports games such as basketball and soccer. 
In demonstrations of basketball or football games, each game segment corresponds to an \emph{unordered} set of $K$ trajectories, which includes $K-1$ players (i.e., agents), and a ball. 
For example, in the professional basketball games illustrated in Figure~\ref{fig:basketballtrajectory}, different players play in different roles (e.g., Shooting Guard) but coordinate together to achieve a common goal (i.e., win the game). 
Roles may change during a game, and the roles of the agents and role assignment mechanism are unknown (to be predicted) at the demonstrations.
Previous multi-agent trajectory prediction approaches~\cite{zheng2016generating,felsen2018will}, use various heuristics (e.g., tree-based role alignment~\cite{sha2017fine}) to assign players to roles, thereby fixing an agent ordering across games. In the most recent research~\cite{yeh2019diverse,sun2019stochastic}, agents are considered as nodes of a fully connected graph, which enables aggregating the players information by applying GNNs. In particular, these methods use reduction operation such as sum or average to aggregate the agents' information. 
However, these methods are simply aggregating the agents information at each time step, and treat the dependency of agents equally. Such dependency plays a key role in the multi-agent games, and the dependencies of different positions/players are significant different. Imaging in a basketball game, a player's future movement does not only depend on the current and past states of other agents, but it also has a strong relation with the dependency of other agents. 
In a soccer game, the right back are less likely to pass the ball to the left midfielder than to the right midfielder. This dependency ignored by previous works needs to be carefully investigated. 

\begin{figure*}[t]
    \centering
    \includegraphics[scale=0.113,trim={0 0 0 0cm},clip]{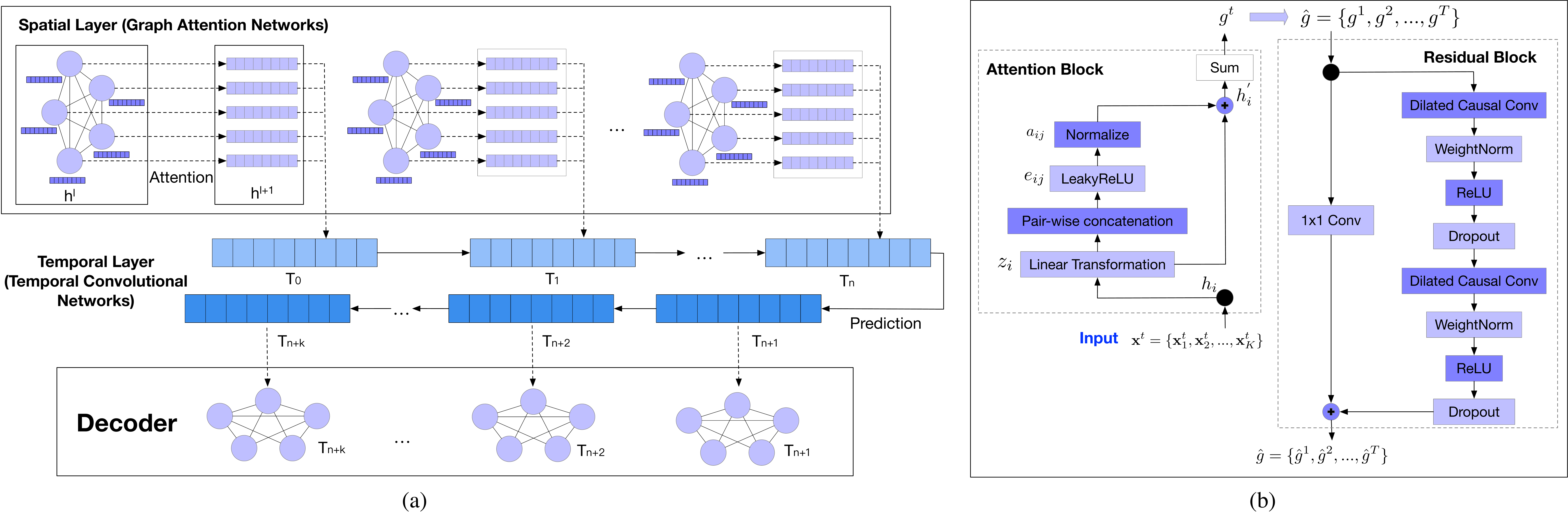}
    \caption{\textbf{Architectural elements in our approach.} (a) \emph{Spatial layer}: at each time step $t$, the players are considered as nodes in a fully connected graph. Hence, we generate a `graph embedding' according to the players information such as their positions at time $t$ by leveraging the attention mechanism that enables modeling the dependencies among players. \emph{Temporal Layer}: We use TCN to model the temporal dependency among the players throughout a demonstration. This convolutional method supports long effective history sizes as well as other features, such as parallelism and stable gradients. (b) \emph{Attention block}: The inputs to the attention block are the tracking data of $K$ players at a time $t$, $\mathbf{x}^t = \{\mathbf{x}^t_1, \mathbf{x}^t_2, . . . , \mathbf{x}^t_K\}$. The output of attention block is the `graph' embedding $g^t$ in the time $t$. Combining with the `graph' embeddings in other time steps, we have $g = {g^1, g^2, ..., g^T}$, which is the input to the residual block in TCN. \emph{TCN Residual block} allows layers to model modifications to the identity mapping, which has repeatedly been shown to benefit very deep networks~\cite{bai2018empirical}. \label{fig:systemoverview}}
\end{figure*}

To address such challenges, in this paper, we propose an attention-based multi-agents trajectory prediction model. To the best of our knowledge, we are the first to propose an approach that is able to capture the fine-grained dependency among the players. Our model can deliver improved forecast to current trajectories prediction approach by:
\begin{itemize}
\item Using fully-connected graph structure to achieve permutation equivariant.
\item Introducing the attention mechanism. The attention coefficients enable modeling dependencies among agents. We are the first to leverage attention mechanism for modeling multi-agents trajectory prediction.
\item Using a temporal convolutional network (TCN) as the sequential model (instead of recurrent networks such as LSTM) to support long effective history and provide important features such as parallelism and stable gradients. We are also the first to use the convolutional sequential model for multi-agents trajectory prediction.
\item The proposed approach is trained end-to-end to predict future states of agents.
\end{itemize}


To verify the performance of the proposed model, we have tested it on two different datasets: basketball trajectories (from Toronto Raptors NBA games illustrated by Figure~\ref{fig:basketballtrajectory}), and soccer trajectories (from 45 games in the Premier League). We show that our approach is able to predict the future trajectories more accurately than the state-of-the-art. 

\paragraph*{Organization} The rest of the paper is organized as follows. 
Section~\ref{sec:ourmethod} describes the main design of our model. Section~\ref{sec:Experimentsetup} shows the baseline models we compare in the experiments, as well as the datasets and how we preprocess the datasets. We report the evaluation results in Section~\ref{sec:evaluation}. Finally, we conclude the paper in Section~\ref{sec:Conclusion}.

\section{Our Method}  \label{sec:ourmethod}

In this section, we present the architecture of our model. In particular, we first discuss the problem setting in this paper. Then, we introduce the architecture of proposed approach (the overview of the architecture is shown in Figure~\ref{fig:systemoverview}) that includes a spatial module (graph structure with attention mechanism) and a temporal module (TCN). Last, we show how our models is trained end-to-end to infer the future trajectories of agents.

\subsection{Problem Setting}

In our problem setting, we assume that we have $K$ agents (i.e., K-1 players, and a ball in a sports game scenario) with a common goal (or sequence of goals) in a set of demonstrations, $D=\{D_1, D_2, ..., D_U\}$.
Let $\mathbf{x}_k^t \in \mathbb{R}^2$ denote the 2-dimensional location of agent $k$ at time $t$, and  $\mathbf{x}^t = \{x^t_1, x^t_2, . . . , x^t_K\}$ be the corresponding states (e.g., positions in sports game) of $K$ agents at time $t$. Consequently, $D_i = \{x^1, x^2, . . . , x^T\}$ is a demonstration that includes a set of `snapshots' of agents' states, corresponding to one segment of a game, where $T$ is the total length in this segment, and $D$ represents all demonstration from the data. Our goal is to build an end-to-end approach to model/predict the trajectories of agents over the sequential data $D$.

\subsection{Spatial Module with Attention Mechanism}

Trajectory data consists of unordered $K$ trajectories, and do not contain the information of the order of such $K$ agents. In order to build a model where the modeled probability is the same across demonstrations in $D$, we need to handle this ordering discrepancy. Prior works \cite{zheng2016generating,felsen2018will,sha2017fine,lucey2013representing} solve this ordering discrepancy problem by `sorting' the agents before (or during) modeling. In other words, similar behaving agents are sorted at the same index across different demonstrations. In this paper, to solve the ordering discrepancy problem, we use a graph-based method to achieve permutation equivariant. In particular, $K$ agents are considered as $K$ nodes that are fully connected in a graph, and the agents' states are the attributes of the corresponding nodes. Hence, a demonstration $D_i$ can be considered as a graph that are dynamically changing the nodes' attributes over the time. We then use a graph-based neural network that handles the graph's input properly because they propagate on each node respectively, ignoring the input order of nodes. In other words, the output of the spatial module is invariant for the order the agents. 

In addition, we also introduce the attention mechanism to the spatial module. Attention-based graph model are widely discussed in several research areas such as traffic prediction~\cite{guo2019attention,yu2017spatio}, action recognition~\cite{yan2018spatial}, multivariate time series modeling~\cite{wu2020connecting}, and code similarity detection~\cite{buggraph}.
Here our approach is able to model the fine-grained dependency among the players across the demonstrations. Our approach computes the hidden representations of each node in the graph, by attending over its neighbors, following a self-attention strategy. Comparing with the vanilla GNN-based methods \cite{sun2019stochastic,yeh2019diverse}, our attention-based approach has the following benefits:

\begin{itemize}
  \item It employs the attention coefficient that is able to model the fine-grained dependency among the players. For example, the coefficient between node $i$ and node $j$ indicates the importance of node $j$'s features to node $i$, which reveals the relation/dependency between player $i$ and player $j$.
  \item The operation of our approach is efficient. We follow the design of Graph Attention Network (GAT) and it is parallelizable across node neighbor pairs. Hence our model allows for (implicitly) assigning different importances to nodes of a same neighborhood, enabling a leap in model capacity~\cite{velivckovic2017graph}. 
\end{itemize}

A single round graph convolution with attention mechanism is illustrated in Figure~\ref{fig:systemoverview}(b). In particular, we generate a graph embedding $g^t$ at time $t$. The output of the spatial module for a demonstration is a set of graph embeddings $\mathbf{g} = \{g^1, g^2, ..., g^t\}$, given an known input sequence $D_i = \{\mathbf{x}^1, \mathbf{x}^2, . . . , \mathbf{x}^t\}$. Detailed information of spatial module is discussed in APPENDIX.A.

\subsection{Temporal Module with Convolutional Nets}

Instead of using recurrent networks (e.g., LSTM) to model the temporal dependency in a demonstration $D_i$, our temporal module is built on a convolutional network --- TCN, that employs very deep networks (augmented with residual layers) and dilated convolutions to support long effective history sizes. In this paper, we use TCN as the sequential model to predict the trajectories. In addition to supporting long effective history, TCN also has the following advantages: 1) \emph{Parallelism}: unlike in RNNs where the predictions for later timesteps must wait for their predecessors to complete, convolutions can be done in parallel. 2) \emph{Stable gradients}: unlike recurrent architectures, TCN has a backpropagation path different from the temporal direction of the sequence, and thus avoids exploding/vanishing gradients. 3) \emph{Low memory requirement for training}: RNNs and LSTMs can easily use up a lot of memory to store the partial results for their multiple cell gates. However, in a TCN the filters are shared across a layer, with the backpropagation path depending only on network depth. 

The goal of the temporal module is to model a sequence of graph embeddings $\mathbf{g} = \{g^1, g^2, ..., g^t\}$ generated from the spatial module, and infer the future unseen states of the agents $\mathbf{x} = \{x^{t+1}, x^{t+2}, ..., x^T\}$, where $\mathbf{g} \in \mathbb{R}$. Note, the key constraint in this module is that to predict the output $x^{t+m}$ for time $t+m$, we are only able to use those inputs that have been previously observed: $\{g^1, g^2, ..., g^t, ..., g^{t+m-1}\}$.

By applying the temporal module $T$ on $\mathbf{g} = \{g^1, g^2, ..., g^t\}$, we have the predicted states of the graphs at unseen time steps $t+1, t+2, ..., T$, as shown in Equation~\ref{eq:temporalprediction}:
\begin{equation}
  \mathbf{\hat{g}} = \{\hat{g}^{t+1}, \hat{g}^{t+2}, ..., \hat{g}^{T}\} = T(\mathbf{g})
  \label{eq:temporalprediction}
\end{equation}

Detailed information of temporal module is discussed in APPENDIX.B.

\subsection{End-to-end training}

In order to compare with the actual agents' 2-D states $\mathbf{x}$, we apply a fully connected layer to $\mathbf{\hat{g}}$ to reduce its dimension:
\begin{equation}\label{eq:reducedemention}
  \mathbf{\hat{x}} = \{\hat{x}^{t+1}, \hat{x}^{t+2}, ..., \hat{x}^{T}\} = F(\mathbf{\hat{g}})
\end{equation}
Then the goal is to minimize the $\ell_2$ loss between $\mathbf{\hat{x}} = \{\hat{x}^{t+1}, \hat{x}^{t+2}, ..., \hat{x}^{T}\}$ and the agents' actual states $\mathbf{x} = \{x^{t+1}, x^{t+2}, ..., x^{T}\}$:
\begin{equation}\label{eq:Loss}
L_{2}(\mathbf{x}, \mathbf{\hat{x}})
\end{equation}
Here $\mathbf{x}$ includes a set of states $x_k^{t+m}$ that are the positions of agent $k$ at time $t+m$. 

Note that it is not a one-time feed forward prediction for all unseen time steps. To predict a distant positions (e.g., $x^{t+i}$), we first decode all its previous positions, $\{x^{t+1}, x^{t+2}, ..., x^{t+i-1}\}$, to guarantee the accuracy of the prediction.



\section{Experiments Setup} \label{sec:Experimentsetup}
In this section, we discuss the experiment setup, including the baseline models we compare in the evaluation, datasets we use, and the implementation details.

\subsection{Models}\label{sec:models}
We compare our proposed approach with several deep learning baselines:
\begin{itemize}
\item \textbf{Velocity}: We use the velocity inference as a simple baseline for the sanity check, i.e., each of the agent's predictions is linearly inferred using its past observed velocity.
\item \textbf{LSTM~\cite{hochreiter1997long}} is recurrent sequential model baseline implemented using a set of gates to control the flow of information. The model uses an MLP as decoder for prediction.
\item \textbf{TCN~\cite{bai2018empirical}} is convolutional sequential model baseline that can provide several advantages than recurrent sequential models. The model uses an MLP as decoder for prediction.
\item \textbf{HMM + LSTM~\cite{le2017coordinated}} is a permutation non-equivalent model that leverage HMMs to capture the orders of agents, and use LSTM to model the temporal dependency.
\item \textbf{Graph + VRNNs~\cite{yeh2019diverse}} is a graph-based permutation equivalent models. It leverage vanilla GNNs for permutation equivalent, and VRNNs for temporal dependency.
\item \textbf{Graph + Attention + TCN (Ours)} is a fully connected graph modeled by attention mechanism to capture the interactions among agents. TCN is used for modeling the temporal dependency to avoid the drawbacks raised by recurrent models.
\item \textbf{Other Variants}: We also consider the models with different components of the neural networks (i.e., \textbf{Graph + Attention + LSTM} and \textbf{HMM + TCN}) to verify the necessity of each component.
\end{itemize}

\subsection{Datasets}\label{sec:datasets}
In this work, we use two datasets in the experiments:
\begin{itemize}
    \item \textbf{Basketball dataset}\footnote{NBA Toronto Raptors Game Records: \url{http://www.cs.toronto.edu/~urtasun/courses/CSC2541_Winter17/project_2.pdf} (Feb, 2017)} is provided with Sportvue trajectory and play-by-play data for 42 Toronto Raptors games in Fall 2015. We only use the trajectory data in our experiment. The original dataset includes 10 players and 1 ball, and records the trajectories at 25 frames per second. It covers the whole court throughout the game. For each frame, the position of the players and the ball are provided. We preprocess the data by the following steps: 1) removing the timeframe where the ball or players are not presented on court; 2) Discarding the timeframes that are not single sided; 3) resampling the dataset to 5Hz; and 4) normalizing the trajectories to be in the range of [-1, 1]. 
    \item \textbf{Soccer dataset~\cite{le2017coordinated}} contains trajectories of 22 soccer players and 1 ball from multiple anonymous teams in the Premier League. The dataset amounts to equivalently and approximately 45 games worth of playing time, with redundant and ``dead'' situations removed. It includes a training set with 7,500 sequences and two separate sets of sequences for testing. The sequences are in different length and they are sampled at 10Hz. We follow the work~\cite{yeh2019diverse} to preprocess the data. First we split the data into segments of length 50 by using a sliding window with 50\% overlap on both the training and test set. The trajectories are centered and normalized to be in the range of [-1, 1]. Note that, although the goal keepers tend not to move much in a demonstration, we still model them because that we believe the movement of goal keepers would affect other players' trajectories. 
\end{itemize}







\setlength\extrarowheight{2pt}
\begin{table*}[!htbp]
  \centering
    \caption{Quantitative results on the \emph{Basketball} and \emph{Soccer} datasets, modeling offense team and defense team respectively. Each demonstration includes 10 seconds in total, where 6-second data observed, and 4-second data to be predicted. We report mean and standard deviation of the mean. Lower numbers are better, and bold is the best results.  \emph{Average error} and \emph{Max error} have the units of feet (in the basketball dataset) or meter (in the soccer dataset); and \emph{miss rate} has the units of percentage.}
    \setlength\tabcolsep{3.5pt} 
    \small
    \begin{tabular}{l|c |c|c c c|c c c}
      \Xhline{2\arrayrulewidth}
      \multirow{2}{2.8cm}{\centering \textbf{Methods}}  &\multirow{2}{*}{\textbf{Order}} & \multirow{2}{*}{\textbf{Team}}  & \multicolumn{3}{c|}{\textbf{\emph{Basketball} (ft)}} & \multicolumn{3}{c}{\textbf{\emph{Soccer} (m)}} \\ \cline{4-9} 
           &   &   & Avg $\ell_2$ error   & Max error & Miss rate &Avg $\ell_2$ error   & Max error  & Miss rate \\ \midrule
  Velocity          &  None  & \multirow{8}{*}{Defense}  &$09.80 \pm .03$        &  $16.10 \pm .05 $       &  $73.01 \pm .09 $     &   $04.63 \pm .02 $      &  $08.84 \pm .02 $    &   $81.41 \pm .12 $  \\ 
  LSTM          &  None  &                               &$10.02 \pm .04$        &  $18.33 \pm .05 $       &  $75.52 \pm .09 $     &   $05.37 \pm .02 $      &  $09.57 \pm .02 $    &   $84.21 \pm .12 $  \\ 
  TCN          &  None  &                                &$09.42 \pm .02$        &  $15.75 \pm .04 $       &  $72.74 \pm .09 $     &   $04.17 \pm .02 $      &  $07.94 \pm .02 $    &   $77.51 \pm .12 $  \\ 
   
  HMM + LSTM           &    Role-based     &           &$09.24 \pm .03$        &  $13.59 \pm .04$        &  $70.74 \pm .08 $     &   $03.34 \pm .01 $      &  $07.67 \pm .02 $    &   $68.46 \pm .07 $  \\
   HMM + TCN  &    Role-based      &                   &$08.79 \pm .02$        &  $12.14 \pm .04$        &  $67.81 \pm .08 $     &   $03.08 \pm .01 $      &  $06.99 \pm .02 $    &   $64.21 \pm .07 $  \\   
   Graph + VRNNs&   Equivariant     &                  &$07.22 \pm .01$        &  $\pmb{9.88 \pm .02} $       &   $63.21\pm.05$       &   $03.02 \pm .01 $      &  $06.21 \pm .01 $    &   $62.11 \pm .04 $ \\  
   Graph + Attention + LSTM      &   Equivariant&      & $07.21 \pm .01$       & $10.21 \pm .02$          &  $64.64 \pm .05$     &    $02.91 \pm .01$       &  $06.06 \pm .02$      &  $58.75 \pm .02$     \\ 
   Graph + Attention + TCN  &   Equivariant    &      & $\pmb{06.95 \pm .01}$ &   $09.91 \pm .01$ & $\pmb{59.44 \pm .05}$ &   $\pmb{02.65 \pm .01}$ &  $\pmb{05.81 \pm .02}$    &  $\pmb{55.63 \pm .03}$     \\ \midrule

   Velocity     & None  & \multirow{8}{*}{Offense}     &  $10.21 \pm .04 $     &   $16.50 \pm .05 $      &  $74.23 \pm .08 $     & $04.67 \pm .02 $        &  $08.96 \pm .02 $         &   $83.02 \pm .11 $    \\ 
   LSTM          &  None  &                            &$10.44 \pm .03$        &  $18.94 \pm .05 $       &  $76.23 \pm .09 $     &   $05.89 \pm .02 $      &  $10.40 \pm .02 $    &   $85.13 \pm .13 $  \\ 
   TCN          &  None  &                             &$09.94 \pm .03$        &  $16.22 \pm .05 $       &  $73.58 \pm .09 $     &   $04.84 \pm .02 $      &  $08.75 \pm .02 $    &   $80.14 \pm .12 $  \\ 
  
   HMM + LSTM  &   Role-based       &                  & $09.51 \pm .03 $      &   $14.23 \pm .03$       &  $71.33 \pm .07$      &  $03.78 \pm .01 $       &  $07.82 \pm .02 $       &    $70.22 \pm .05 $   \\  
   HMM + TCN  &    Role-based     &                   &$09.23 \pm .03$        &  $13.59 \pm .04$        &  $70.74 \pm .08 $     &   $03.34 \pm .01 $      &  $07.68 \pm .02 $    &   $68.41 \pm .07 $  \\ 
   Graph + VRNNs  &   Equivariant  &               &    $08.80 \pm .01 $   &   $13.56 \pm .04 $      &    $68.71\pm.06$           & $03.17 \pm .01 $        &  $06.50 \pm .01 $       &   $64.47 \pm .05 $   \\ 
  Graph + Attention + LSTM &     Equivariant      &     & $07.68 \pm .01$     & $11.92 \pm .01$        &  $65.33 \pm .05$        &  $\pmb{03.04 \pm .02}$       &  $06.77 \pm .02$           &  $63.10 \pm .03$     \\ 
 Graph + Attention + TCN  &   Equivariant      &     & $\pmb{07.54 \pm .01}$ & $\pmb{10.64 \pm .01}$   &  $\pmb{61.21 \pm .05}$&  $03.08 \pm .01$  &  $\pmb{06.20 \pm .02}$  &  $\pmb{62.10 \pm .03}$     \\ \Xhline{2\arrayrulewidth}
    \end{tabular} \label{tb:basiccomparison}
  \end{table*}

\subsection{Implementation details}

The final goal is to predict the future trajectories of the agents. To do so, according to their past trajectories, we first infer the current state (2-D position) of the all the objects (players and ball). In particular, we minimize normalized $\ell_2$ distance between the ground truth locations and the predicted locations of all objects. Like imitation learning~\cite{le2017coordinated}, we predict the future trajectories also by minimizing $\ell_2$ loss.
We train all the models using the standard Adam optimizer. To prevent over-fitting, we select the best performing model using log-likelihood on the validation set.  The models are trained on 4 V100 GPUs with synchronous training with batch size of 8 per GPU. The initial learning rate is 0.0005. The learning rate is decayed exponentially by a factor of 0.999 per epoch.

\section{Evaluation}\label{sec:evaluation}
In this section, we compare our approach with various baselines (discussed in Section~\ref{sec:models}) on two datasets: modeling of basketball and soccer game trajectories.

\subsection{Evaluation Metrics}
We evaluate the models on the task of predicting future trajectories, i.e., conditioned on the first $n$ seconds of all agents' trajectories, we predict the following (future) $m$ seconds trajectories. To demonstrate the efficiency and effectiveness of our approach, we evaluate on the following metrics:
\begin{itemize}
    \item \textbf{Average error} is the $\ell_2$ error between predicted trajectories and the ground truth, averaged over each time step for each agent (as shown in Equation~\ref{eq:Loss}). For each test run, we randomly sample 20 data points and report the \emph{Average error}.
    \item \textbf{Max error} is the maximum $\ell_2$ error between the prediction and ground truth for an agent trajectory, averaged over all agent trajectories. For each test run, we randomly sample 20 data points and report the \emph{Max error}.
    \item \textbf{Miss rate} is calculated as the fraction of time the $\ell_2$ error exceeds 3 ft in basketball game, 1 meter in soccer game. This is reported on the best out of 20 data points per test run.
\end{itemize}

\subsection{Basic comparison}
We first compare our approach with the baselines in \emph{average error}, \emph{max error} and \emph{miss rate}. In particular, we compare the methods run on offense team as well as the defense team. Note that in the soccer dataset, although the trajectories of goalkeepers are not predicted, the model is conditioned with goalkeeper's information. For the basketball and soccer games, we consider 10 seconds in total, where 6-second data observed, and 4-second data unobserved (to be predicted). Table~\ref{tb:basiccomparison} shows the quantitative results. 

Intuitively, predicting the trajectories of defense players is more accurate than predicting offense players, because the defense players react the actions of offense team. Hence, the offense trajectories contains more straightforward information comparing with the defense motion. This assumption is proved by the evaluation results (Table~\ref{tb:basiccomparison}) where the prediction of defense team improves around 8\% comparing with the offense team. It is observed that the velocity baseline outperforms the simple LSTM in the three metrics. In addition, similar as the results shown in~\cite{bai2018empirical}, convolutional sequential model performs better than the recurrent sequential model (i.e., LSTM). The similar conclusion could also be found in the comparison where the player's order is considered (i.e., role-based, permutation equivariant). 

We also perform several ablation studies to verify the effectiveness of each the components. As the results shown in Table~\ref{tb:basiccomparison}, basically, graph-based models outperform all of the non-graph-based models. Furthermore, the attention mechanism indeed learns varied dependency among agents, as the most of the metrics are lower than the non-attention graph-based model. Note that, the datasets we used in this simulation do not differentiate the teams in the offense or defense teams. In another word, the offense model (or defense model) learns the behavior of multiple offense teams (or defense teams), and it does not represent for the strategy of a single team. As we have discussed in the Section~\ref{sec:Introduction}, the attention mechanism is able to learn the patterns (or strategy) of a specific team. In the next section, the evaluation is performed on two datasets divided by the basketball dataset: \emph{basketball\_tor\_offense}: Toronto Raptors as the offense team, and \emph{basketball\_tor\_defense}: Toronto Raptors as the defense team.

\setlength\extrarowheight{2pt}
\begin{table*}[!htbp]
  \centering
    \caption{Quantitative results on \emph{basketball\_tor\_offense} and \emph{basketball\_tor\_defense} datasets, modeling offense team and defense team respectively. Each demonstration includes 10 seconds in total, where 6-second data observed, and 4-second data to be predicted. We report mean and standard deviation of the mean. Lower numbers are better, and bold is the best results.  \emph{Average error} and \emph{Max error} have the units of feet (in the basketball dataset) or meter (in the soccer dataset); and \emph{miss rate} has the units of percentage.}
    \setlength\tabcolsep{3.5pt} 
    \small
    \begin{tabular}{l|c |c|c c c|c c c}
      \Xhline{2\arrayrulewidth}
      \multirow{2}{2.8cm}{\centering \textbf{Methods}}  &\multirow{2}{*}{\textbf{Order}} & \multirow{2}{*}{\textbf{Team}}  & \multicolumn{3}{c|}{\textbf{\emph{basketball\_tor\_defense} (ft)}} & \multicolumn{3}{c}{\textbf{\emph{basketball\_tor\_offense} (ft)}} \\ \cline{4-9} 
           &   &   & Avg $\ell_2$ error   & Max error & Miss rate &Avg $\ell_2$ error   & Max error  & Miss rate \\ \midrule
  HMM + LSTM&Role-based&\multirow{4}{*}{Defense}       &$09.24 \pm .03$        &  $14.01 \pm .04$        &  $70.64 \pm .08 $     &$09.23 \pm .03$        &  $14.20 \pm .04$        &  $70.72 \pm .08 $   \\
   Graph + VRNNs&   Equivariant     &                  &$07.02 \pm .01$        &  $09.44 \pm .02 $       &   $63.01\pm.05$     &$07.22 \pm .01$        &  $\pmb{09.80 \pm .02}$       &   $63.21\pm.05$ \\  
   Graph + Attention + LSTM      &   Equivariant&      & $05.41 \pm .01$       & $07.21 \pm .02$          &  $53.64 \pm .05$     & $07.01 \pm .01$       & $10.31 \pm .02$          &  $62.64 \pm .05$     \\ 
   Graph + Attention + TCN  &   Equivariant    &      & $\pmb{04.95 \pm .01}$ &   $\pmb{06.12 \pm .01}$ & $\pmb{48.44 \pm .05}$      & $\pmb{06.20 \pm .01}$ &   $09.81 \pm .01 $ & $\pmb{59.44 \pm .05}$     \\ \midrule

   HMM + LSTM&Role-based&\multirow{4}{*}{Offense}&     $09.41 \pm .03 $      &   $14.25 \pm .03$       &  $71.13 \pm .07$      &$09.33 \pm .03$        &  $14.11 \pm .04$        &  $70.69 \pm .08 $   \\  
   Graph + VRNNs  &   Equivariant  &               &    $08.82 \pm .01 $   &   $13.51 \pm .04 $      &    $68.11\pm.06$           &    $07.62 \pm .01 $   &   $12.11 \pm .04 $      &    $64.31\pm.06$   \\ 
  Graph + Attention + LSTM &     Equivariant      &     & $07.63 \pm .01$     & $11.92 \pm .01$        &  $65.23 \pm .05$        &  $07.52 \pm .02$       &  $10.64 \pm .02$           &  $63.34 \pm .03$     \\ 
 Graph + Attention + TCN  &   Equivariant      &     & $\pmb{07.44 \pm .01}$ & $\pmb{10.54 \pm .01}$   &  $\pmb{61.11 \pm .05}$&  $\pmb{06.21 \pm .01}$  &  $\pmb{09.84 \pm .04}$  &  $\pmb{58.92 \pm .05}$     \\ \Xhline{2\arrayrulewidth}
    \end{tabular} \label{tb:modelingteam}
  \end{table*}
\subsection{Comparison in modeling a specific team}
In this section, we verify the ability of our approach to learn (and infer) the strategy of a specific team. In particular, we compare our approach with the methods \emph{HMM + LSTM}, \emph{Graph + VRNNs} and \emph{Graph + Attention + LSTM}, that have comparable results in the previous section. The quantitative results is shown in Table~\ref{tb:modelingteam}.

The predictions of offense players in the \emph{basketball\_tor\_defense} dataset and defense players in the \emph{basketball\_tor\_offense} dataset have the similar numbers as the predictions in Table~\ref{tb:basiccomparison}. This is because that, the datasets for these evaluations are mixed with several different teams. What the models learned is the general behavior of basketball offense (or defense) players. 

However, this is not the case for the predictions of attention-based approaches when modeling on the datasets include only one team (i.e., Toronto Raptors) --- predicting defense trajectories in the \emph{basketball\_tor\_defense} dataset and predicting offense trajectories in the \emph{basketball\_tor\_offense} dataset. As shown in Table~\ref{tb:modelingteam}, attention-based approaches (i.e., \emph{Graph + Attention + LSTM} and \emph{Graph + Attention + TCN}) largely outperform  other baselines (i.e., \emph{HMM + LSTM}, \emph{Graph + VRNNs}). In particular, for \emph{HMM + LSTM} and \emph{Graph + VRNNs}, performance of predicting a single team is slightly better than the prediction in mixed teams; while for the attention-based methods, modeling in a single team has much lower $\ell_2$ error and miss rate. This proves that our attention based permutation equivariant method has the ability to model a single team and learn the dependency of that team. Namely, our approach learns the strategy of the team.

\section{Conclusion} \label{sec:Conclusion}
We study the problem of multi-agents trajectory prediction, and propose a spatial-temporal trajectory prediction approach. In particular, we use a fully-connected graph structure to achieve permutation equivariant. In addition, the attention mechanism is used for modeling the fine-grained dependency of the agents. In addition, instead of utilizing the recurrent networks (e.g., VRNN, LSTM), our method use a TCN as the sequential model support long effective history and provide important features such as parallelism and stable gradients.
The evaluation shows that our approach is able to predict the future trajectories of sports games more accurately than the state-of-the-art. 

\section*{Acknowledgment}
The authors would like to thank the anonymous reviewers for their suggestions.  This work was supported in part by National Science Foundation grants 1618706 and 1717774.

\bibliographystyle{IEEEtran}

\bibliography{mybib.bib}

\section{APPENDIX}
In Appendix, we show the detailed transformation of the special module and the temporal module.

\subsection{Spatial Module} 
The input of the spatial module is $\mathbf{x}^t = \{x^t_1, x^t_2, . . . , x^t_K\}$ that includes $K$ agents' states. We denote the feature vector of agent $i$ as to $h_i$, hence we have the feature vectors of all the nodes $\mathbf{h} = \{h_1, h_2, ..., h_K\}$. We first apply a linear transformation to $h_i$ as shown in in Equation~\ref{eq:GAT_linear_trans}:
\begin{equation}\label{eq:GAT_linear_trans}
z_i = W\cdot h_i
\end{equation}
where $W$ is a learnable weight matrix. Equation (2) computes a pair-wise un-normalized attention score between every two neighbors (e.g., node $i$ and node $j$)
\begin{equation}\label{eq:GAT_unnormal_attention}
e_{ij} = LeakyReLU(a^T\cdot (z_i \parallel z_j))
\end{equation}
Here, it first concatenates the $z$ embeddings of the two nodes, where $\left |  \right |$ denotes concatenation, then takes a dot product of it and a learnable weight vector $a^T$, and applies a activation function (i.e., \textit{LeakyReLU}) in the end. Note, this suggests that the attention scores can be viewed as edge data, which indicates the dependency between two nodes. Equation~\ref{eq:GAT_normal_attention} applies a softmax to normalize the attention scores on each node's incoming edges. 
\begin{equation}\label{eq:GAT_normal_attention}
a_{ij} = \frac{exp(e_{ij})}{\sum_{k\in K}^{ }exp(e_{ik})}
\end{equation}
Equation~\ref{eq:GAT_node_embedding} is similar to the convolution method, where the embeddings from neighbors are aggregated together to generate a node embedding $h'_i$. The difference is that, in our method, the aggregation is scaled by the attention scores.
\begin{equation}\label{eq:GAT_node_embedding}
h'_i = \sigma \left [ \sum_{j\in K}^{ } a_{ji}^{(l)} \cdot z_j^{(l)} \right ]
\end{equation}
In the end, in Equation~\ref{eq:GAT_graph_embedding}, by summing the node embedding $\mathbf{h} = \{h'_1, h'_2, ..., h'_K\}$, we aggregate the agents information at time $t$, and hence generate the graph embedding $g^t$. 
\begin{equation}\label{eq:GAT_graph_embedding}
g^t = \sum_{i\in K}^{ } h'_{i}
\end{equation}
Attention coefficient is calculated symmetrically between pairs of nodes and the summation operators (i.e., Equation~\ref{eq:GAT_normal_attention},~\ref{eq:GAT_node_embedding} and~\ref{eq:GAT_graph_embedding}) ignores the ordering of the nodes, which enables the permutation equivariance. At the end, the output of the spatial module for a demonstration is a set of graph embeddings $\mathbf{g} = \{g^1, g^2, ..., g^t\}$, given an known input sequence $D_i = \{\mathbf{x}^1, \mathbf{x}^2, . . . , \mathbf{x}^t\}$.

\subsection{Temporal Module}
In order to convolutionally modeling the sequential data, our temporal module leverages \emph{dilated convolution}~\cite{oord2016wavenet}, and \emph{residual connection}~\cite{he2016deep}. By applying dilated convolution, the module is able to support exponentially large receptive field. Considering that we have a kernel (filter) $k: \{0, ..., f-1\} \in \mathbb{R}$, the dilated convolution operation $C$ on $s$th element of the sequence is defined as:
\begin{equation}
C(s) = \sum_{f-1}^{i=0}k(i) \cdot \mathbf{g}_{s-d\cdot i} 
\end{equation}
where $d$ is the dilation factor, $f$ is the kernel size, and $s - d \cdot i$ indicates the indexes of the past elements that convolutionally aggregated. Dilation is thus equivalent to introducing a fixed step between every two adjacent filter taps. When $d = 1$, a dilated convolution reduces to a regular convolution. Note, the effective history in one convolutional layer is $(k-1)*d$. Therefore, there are two ways to increase the receptive field: employing larger filter size $k$, or increasing the dilation factor $d$.

Since the receptive field of temporal module not only depends on filter size $k$ and dilation factor $d$, it also depends one the network depth $n$. Therefore, stabilization of a deeper and larger TCNs is essential. To support a deeper network, we follow~\cite{bai2018empirical} to use a residual block that effectively allows layers to learn modifications to the identity mapping rather than the entire transformation, which has repeatedly been shown to benefit very deep networks. A residual block contains a branch leading out to a series of transformations $R$, whose outputs are added to the input of the block.
\begin{equation}\label{eq:residual_block}
  o = Activation(x + R(x))
  \end{equation}
As illustrated in Figure~\ref{fig:systemoverview}.(b), within a residual block, we have
\begin{itemize}
  \item two layers of dilated causal convolution and non-linearity (i.e., ReLU~\cite{nair2010rectified})
  \item a weight normalization~\cite{salimans2016weight} to the convolutional filters.
  \item a spatial dropout~\cite{srivastava2014dropout} was added after each dilated convolution for regularization.
\end{itemize}
Note, the input and output of residual block have different widths, so we use an additional $1*1$ convolution to ensure that the operation $\oplus$ is on the same shape tensors. By applying the temporal module $T$ on $\mathbf{g} = \{g^1, g^2, ..., g^t\}$, we have the predicted states of the graphs at unseen time steps $t+1, t+2, ..., T$, as shown in Equation~\ref{eq:temproalprediction}:
\begin{equation}\label{eq:temproalprediction}
  \mathbf{\hat{g}} = \{\hat{g}^{t+1}, \hat{g}^{t+2}, ..., \hat{g}^{T}\} = T(\mathbf{g})
\end{equation}

\end{document}